\documentclass[10pt,twocolumn,letterpaper]{article}

\usepackage{iccv}
\usepackage{times}
\usepackage{epsfig}
\usepackage{graphicx}
\usepackage{amsmath}
\usepackage{amssymb}

\usepackage{multirow}
\usepackage{xcolor}
\usepackage{subcaption}

\usepackage[pagebackref=true,breaklinks=true,letterpaper=true,colorlinks,bookmarks=false]{hyperref}

\iccvfinalcopy 

\def\iccvPaperID{1549} 

\ificcvfinal\fi
\begin{document}

\title{On the Role of Geometry in Geo-Localization}

\author{Moti Kadosh\\
Tel-Aviv University\\
Tel-Aviv, Israel\\
{\tt\small moti.kadosh84@gmail.com}
\and
Yael Moses\\
Interdisciplinary Center Herzliya\\
Herzliya, Israel\\
{\tt\small yael@idc.ac.il}
\and
Ariel Shamir\\
Interdisciplinary Center Herzliya\\
Herzliya, Israel\\
{\tt\small arik@idc.ac.il}
} 

\maketitle

\begin{abstract}
   Humans can build a mental map of a geographical area to find their way and recognize places. The basic task we consider is geo-localization -- finding the pose (position \& orientation) of a camera in a large 3D scene from a single image. 
   We aim to experimentally explore the role of geometry in geo-localization in a convolutional neural network (CNN) solution. We do so by ignoring the often available texture of the scene.
   We therefore deliberately avoid using texture or rich geometric details and use images projected from a simple 3D model of a city, which we term \emph{lean images}. Lean images contain solely information that relates to the geometry of the area viewed (edges, faces, or relative depth).
   We find that the network is capable of estimating the camera pose from the lean images, and it does so  not by memorization but by some measure of geometric learning of the geographical area.
   The main contributions of this paper are: (i) providing insight into the role of geometry in the CNN learning process; and (ii) demonstrating the power of CNNs for recovering camera pose using lean images.
\end{abstract}

\section{Introduction}
\label{introduction}

Recently, several works in the field focused on trying to understand how neural networks represent data and tackle their limits \cite{zhang2016understanding}.
Our paper's main goal is to study the role of geometry in a CNN solution to the geo-localization task rather than propose a working system for application purposes.

What is the geo-localization task? Imagine you are brought blindfolded to a street corner of a city you know well. Now, you remove the blindfold. Can you tell where you are? In the computer vision field, this amounts to estimating the position (and sometimes the orientation) of a camera given its current view.
Although localization devices such as Global Positioning Systems (GPS) have improved significantly over the last years, they often do not work well in city scenes and do not provide highly accurate results. Autonomous cars, drones, and IOT devices are expected to benefit tremendously from the ability to determine their pose (position \& orientation) accurately in their environment.

\begin{figure}[t]
  \centering
   \includegraphics[width=1.0\linewidth]{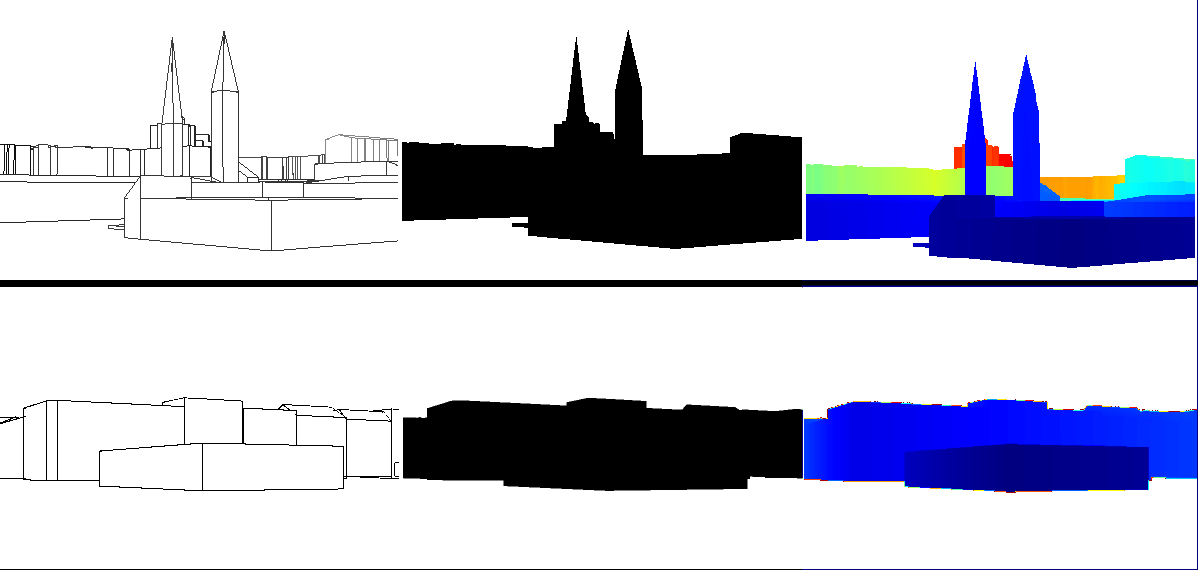} \includegraphics[width=0.8\linewidth]{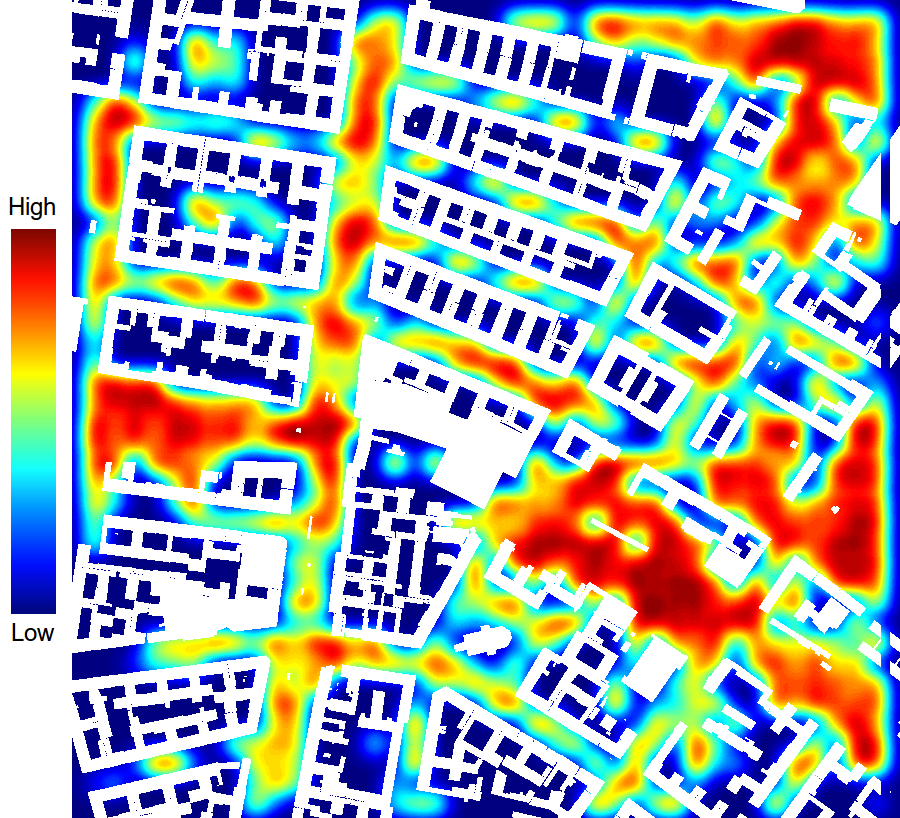}
   \caption{Top: \emph{lean images} contain mostly geometric features: edges (left), faces (center), and depth information (right). We train a CNN to solve the localization problem using such images alone. 
   Bottom: a top view of a city area (buildings are marked as white) where color indicates the localization success rate of the network from red (high) to blue (low). For instance, note how open spaces are more distinct than narrow streets. }
  \label{fig:input_images_and_test_vis}
\end{figure}


\begin{figure}[t!]
  \centering
  \includegraphics[width=1.0\linewidth]{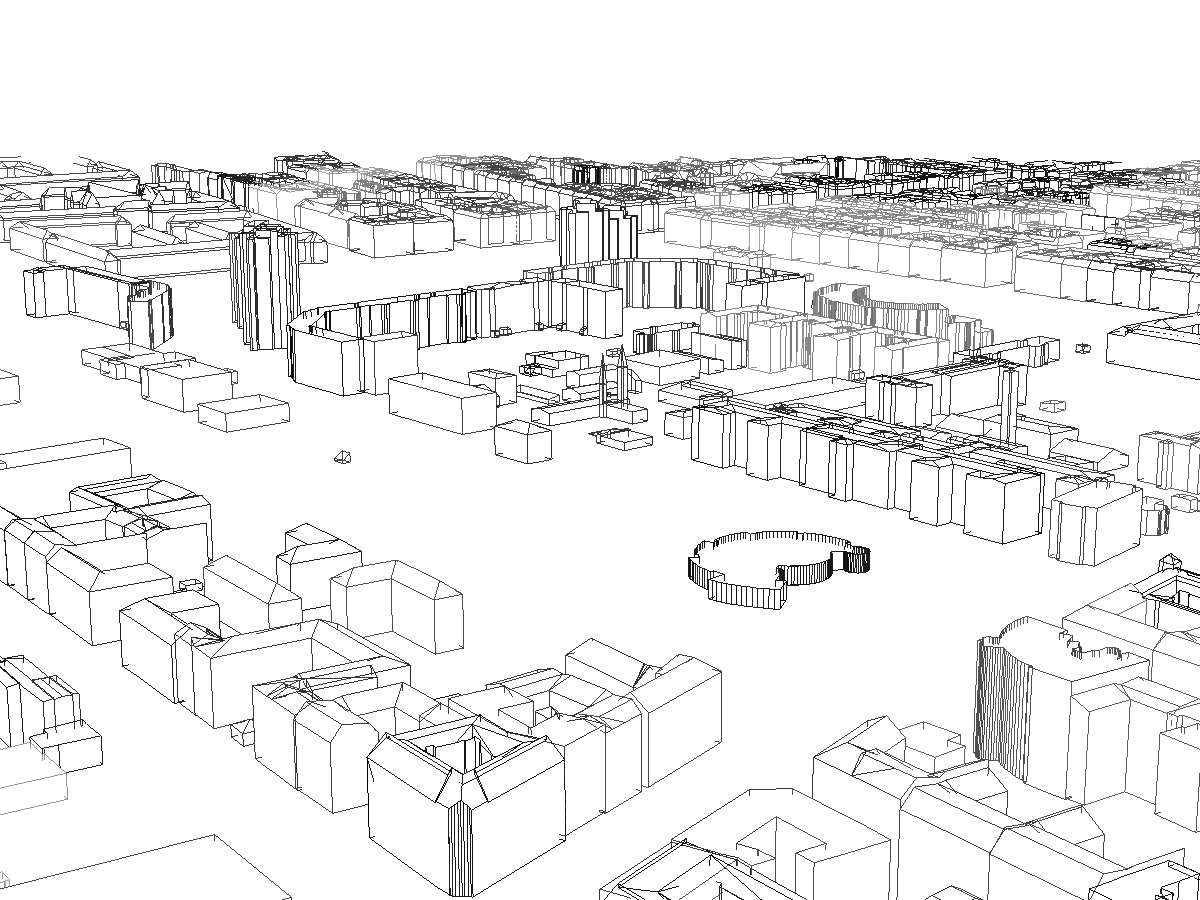}
  \caption{Bird's-eye view of one of the areas we used.}
  \label{fig:high_view}
\end{figure}

A solution for geo-localization, either by a human or by a machine, can use appearance cues (\eg, texture of a unique building), geometric cues (\eg, a unique shape of a building), or both. In the `80s and `90s, many computer vision tasks were based on edge images (\ie, mostly geometry). More recently, significant improvements were obtained in object recognition, scene recognition, and localization tasks, largely by exploiting the appearance of the scene (\eg, color and texture and image features such as SIFT \cite{Stephen2002MobileRobotLocalization, Lowe2004SIFT, li2010location}). Later, these methods were improved by adding coarse geometric constraints to the image features (\eg, \cite{ramalingam2011pose, bansal2014geometric}). Nowadays, methods are based on machine learning, in particular convolutional neural networks (CNNs), where the input is the unprocessed image. Clearly, both appearance and geometry play an important role in these methods.

We aim to explore the role of geometry alone in geo-localization using end-to-end deep learning neural network, while ignoring the often available texture of the scene.
To do this, we consider the geo-localization task using lean images. Lean images contain mostly information that relates to the geometry of the scene while lacking texture or rich geometric details. In particular, we use a city scene and consider two types of binary images that consist of the edges of the buildings' outline and the buildings' facades. In addition, we also consider depth images that contain more geometric information.

Examples of the three types of lean images are shown in Figure~\ref{fig:input_images_and_test_vis} (top). Note that in the first row, the view contains dominant landmarks, while the second row shows very little distinct information that might be expected to assist localization.
Such non-distinct views are very common in large environments such as a city, making localization with lean images very challenging.  Further note that we deliberately do not consider real images or synthetic images with texture, since our goal is to study only the information available from purely geometry information.

We use an untextured 3D mesh model of Berlin \cite{BerlinModel} to generate our data. A bird's-eye view of one of the areas is shown in Figure~\ref{fig:high_view}. 
Using such a model allows us to study the role of geometry for geo-localization in a controlled manner and in a larger scale than ever before, both in terms of the area covered (many city streets) and in terms of the number of images (up to hundreds of thousands).
Our images are obtained simply by projecting the model onto various positions in the scene.
Each image is defined by four parameters: $(x, y)$ represents the camera position on the ground plane and $(\theta, \phi)$ represents the $Yaw$ and $Pitch$ angles of the camera respectively. We assume for simplicity that the picture is taken at a fixed height, and the roll angle is fixed as horizontal.

A typical geo-localization solution will return either the pose from which an image was taken or the most similar image from a database.
We consider two variants of the geo-localization tasks. The first task is recognizing a previously seen view of the scene, which we refer to as \emph{Geo-Matching}. The second is determining the camera pose in a previously unseen view, which we refer to as \emph{Geo-Interpolation}.

The geo-matching task can be regarded as image retrieval from a database of all available views of the city. A na\"{i}ve solution would store all images and then perform a brute-force search in the database. However, this is inefficient and can become infeasible as the database gets larger. Defining a compact representation and an efficient search is clearly desired, and it is often performed by manually engineered image representation (\eg, a dictionary of image features) and an image retrieval approach, including the metric between the stored representation and a target one (\eg, \cite{SivicAndZissermanVideoGoogle, Lowe2004SIFT, robertson2004image, Zhang2006, Nister2006, Schindler2007, Hays2008im2gps, li2010location}).
Neural networks were shown to be effective in geo-localization tasks (\eg, \cite{kendall2015posenet, walch2017image, melekhov2017image}). They may be used  to perform both functions: provide storage and learn the metric. The questions we address here are (i)~whether CNNs can also be used to solve the geo-matching task from lean images and (ii)~whether geometric information is used by the CNN to solve the task.
In the geo-interpolation task the image query is not part of the training set. In this case we ask (iii)~can the CNN generalize and support geo-interpolation in such large environments using only geometric and spatial data?

As discussed in the results section (Sec.~\ref{sec:experiments_and_results}), we found positive answers to all these questions, but the results depend on the number of images and their sampling density. We believe this indicates that networks can learn some sort of a  spatial map for an area using only geometric data, since no colors or textures are available in our data. The success of geo-localization also depends on the specific position. Figure~\ref{fig:input_images_and_test_vis} (bottom) shows how certain positions in the streets of a city are more recognizable than others. 

The paper presents an empirical study regarding the information that can be used by CNNs; we do not propose a practical solution based on lean images. The main contributions of our study are: 
(i) proposing a systematic method to study the role of geometry in CNNs when trained to solve geo-localization tasks; (ii) demonstrating the power of CNNs to use the geometric information efficiently; and (iii) showing that lean images contain sufficient information for solving the geo-matching and geo-interpolation tasks.


\section{Related Work}
\label{sec:related_work}



Place recognition (\eg, recognizing the Eiffel Tower in an image) can be regarded as a coarse geo-localization task. Finding images of the same place is a basic tool for solving this task. Classic approaches use visual features to represent each image in a set of images of a given location (\eg, by a bag of words) and then match a target image with the stored representations (\eg, \cite{Stephen2002MobileRobotLocalization, SivicAndZissermanVideoGoogle, Lowe2004SIFT, robertson2004image}).
Hays \& Efros~\cite{Hays2008im2gps} were the first to address the place recognition task using millions of geo-tagged images. In their study they compare the results obtained when various visual features are used (tiny images, color histograms, texton histograms, line features, gist descriptors with color and geometric context).



In our study, we consider the geo-localization task, where both position and orientation of a camera with respect to a scene should be estimated.
A possible solution can be obtained using triangulation with images with known pose (\eg, \cite{Zhang2006}).
In most studies, 3D models of the scene are used by means of point-clouds (\eg, \cite{Irschara2009, Sattler2011FastImageBasedLocalization, Image2LIDARmatch2013, svarm2014accurate}), Digital Elevation Maps (DEM) (\eg, \cite{baatz2012large, bansal2014geometric}), or full 3D models (\eg, \cite{ramalingam2011pose}). One of the main challenges of these works is to develop an efficient computation of 2D to 3D feature matching. The matching can then be used to determine the query image pose with respect to the model. Computing the matching requires dealing with a very large search space, and outliers must also be discarded. 
Works that deal with these challenges include classic studies of outliers removal (\eg, \cite{fischler1981random, Haralick1989, svarm2014accurate, Li2009ConsensusSetMaximization, sattler2012improving}). 

New 3D feature representation have also been developed (\eg, \cite{Irschara2009, Sattler2011FastImageBasedLocalization}).
Bansal \& Daniilidis \cite{bansal2014geometric} introduce a feature more closely related to the lean images we consider. It consists of 3D corners and direction vectors extracted from a Digital Elevation Map (DEM) to be matched geometrically to the corners and roof-line edges of buildings visible in a street-level query image.

Efficiency and robustness become even more important when dealing with a city-scale 3D model.
A fast method for inliers detection that enables solving the correspondence problem on such a scale was suggested by Sv\"{a}rm \etal~\cite{svarm2017city}. Recent survey on existing localization methods can be found in \cite{piasco2018survey}.

One of the key ideas that bypasses the challenge of defining an efficient and robust 2D-3D feature matching required by the abovementioned methods is to use an end-to-end CNN solution that performs both feature extraction and matching.
PoseNet \cite{kendall2015posenet} is an impressive CNN based approach for solving the pose of real images.
A dataset of images was used for training Google LeNet \cite{szegedy2015going} where the 6-DoF pose of the camera was used as ground truth. The softmax final layer of Google LeNet, which was used for an object classification task, was replaced by a vector for a regression task. The Google LeNet was pre-trained on the ImageNet dataset for the object recognition challenge \cite{imagenet_cvpr09, ILSVRC15}.
Walch \etal \cite{walch2017image} suggested an improvement to the PoseNet CNN architecture by adding an LSTM, which reduces the dimensionality of the feature vector. Melekhov \etal \cite{melekhov2017image} used ResNet34 \cite{he2016deep}, which uses encoder-decoder structure to improve model accuracy.
Kendall \& Cipolla \cite{kendall2016modelling} improved their earlier work \cite{kendall2015posenet} by applying an uncertainty framework to the CNN pose regressor.
In another work, Kendall \& Cipolla \cite{kendall2017geometric} studied the affect of various loss functions on the result of PoseNet.

In our study we assume a 3D model of a city is given. Our setup is very challenging since the model and the images consists of only coarse 3D structure of the scene without texture for computing image features. On the other hand, our images are noise-free and there are no object-level occlusions such as trees, cars and people. 
Our method uses a CNN in a similar manner to PoseNet. However, we use the ResNet50 architecture, also modified for regression, which produced better results for our task.
We trained our network from scratch since we use lean images, which are projections of an untextured 3D model, \ie using pre-training done on texture images is irrelevant.
In addition, we were not limited by data size, as we projected as many images as we chose.

Most importantly, our goal differs from that of the aforementioned methods: whereas they focus on obtaining a better and faster solution for geo-localization, we focus on trying to understand the role of geometry, alone, in geo-localization, by systematically training and testing the same neural network on controlled datasets.

\section{Data}
\label{sec:data}
``Your network is as good as your data" is a common phrase in the world of neural networks. Our case is no different. In this respect, using a 3D model as the data source is highly advantageous: we can sample as many images as necessary from the 3D model in any position, orientation and resolution.

All images used in our study are projections of a 3D model of Berlin \cite{BerlinModel} without textures. This model is very simple, it contains only the geometry of building walls and rooftops, and does not contain any fine geometric details such as window frames or doors (see bird-eyes view in  Figure~\ref{fig:high_view}). We consider three types of images: edge, face, and depth map, see Figure~\ref{fig:input_images_and_test_vis} (top), and we call them lean images since they contain no texture or structural details.
Face images are actually the buildings' facades.

We generated several image datasets that are sampled uniformly along a 4D grid, where each image is defined by its camera pose. That is, $(x, y, \theta, \phi)$, where $(x, y)$ is the position on the ground plane and $(\theta, \phi)$ is the camera orientation.
The density in the $(x, y)$ domain varies between the datasets but fixed in the $(\theta, \phi)$ domain.
Each set of images is created in a defined area of the city. The number of images in the set is determined by the size of the area and the grid sampling density. The three types of lean images were generated for each sample pose.

\begin{figure}[t!]
  \centering
  \includegraphics[width=0.8\linewidth]{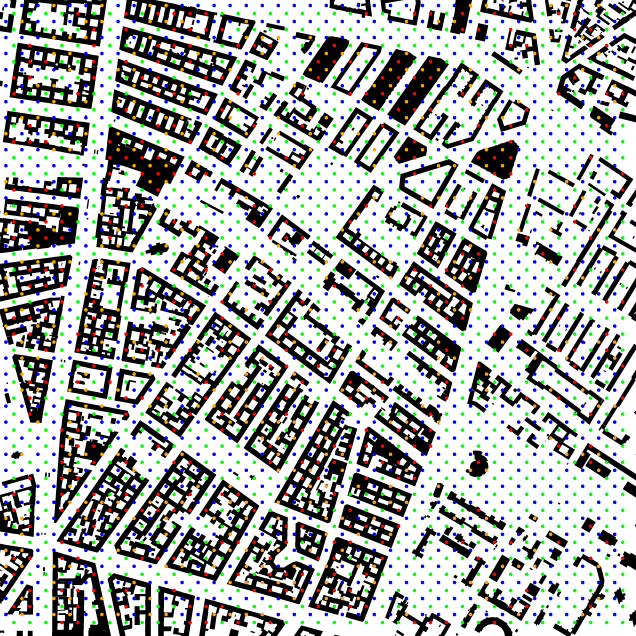}
  \caption{Example of sampling positions on a area of the map. For the training set: green indicates \textcolor{green}{valid} samples and red \textcolor{red}{invalid} samples. For the test set: blue indicates \textcolor{blue}{valid} samples and orange \textcolor{orange}{invalid} samples. (Better viewed on screen.)
  }
  \label{fig:sampling_map}
\end{figure}

When dealing with lean images, care must be taken not to include empty images. For example when the camera is facing a building wall from a short distance. Such images contain almost no visual information and do not contribute to the learning process.
We define an invalid image as an image that has less than $8$~edges, or an image that does not contain a skyline (at least $50\%$ of its top-most pixel row is sky).
Moreover, images associated with a pose inside a building are irrelevant to geo-localization, and are also defined as invalid.
We remove invalid images from both the training and the test sets (see Figure~\ref{fig:sampling_map}).

Although the representation of Euler angles using $(\theta, \phi)$ is natural and easier to comprehend, it suffers from ambiguity and Gimbal lock. Therefore, in practice we use quaternions which offer stability, efficiency and compactness (see~\cite{kendall2015posenet}).
Each image sampled from the 3D model was defined by a 6D pose vector in the form of $(x, y, q_1, q_2, q_3, q_4)$, which in fact represents 4 degrees of freedom.

\section{Network}
\label{sec:network}
We examined several convolutional neural network (CNN) architectures that proved to be successful on object recognition tasks. Specifically, we tested VGG, Google LeNet \cite{szegedy2015going} and ResNet50 \cite{he2016deep}, built for the ImageNet Large Scale Visual Recognition Challenge (ILSVRC) \cite{ILSVRC15, imagenet_cvpr09}.

Our geo-matching task could have been defined as a classification task where each $(x,y,\theta,\phi)$ is considered as a class. However, this would involve learning a huge number of classes $(\sim170K)$.
In addition, a classification setup loses the spatial relations among the images because each class is considered completely unrelated to others. This prevents the network from exploiting the geometric structure and information, and can preclude an answer to one of our main research questions: Can a network use geometric information?

Thus, more suitable for our problem is to consider a CNN for solving a regression task. This also allows to use the same trained CNN for the geo-interpolation task, by directly returning the pose of unseen images in the test set. Otherwise, if a classification CNN was used, it would have required a post-processing of the result, \eg by classic methods such as averaging the $K$-nearest classification matches.
Because the considered CNNs were designed for classification tasks, we follow \cite{kendall2015posenet} and modify them to solve a regression task by simply removing the last softmax layer and replacing it by a fully connected layer of our result vector $(x, y, q_1, q_2, q_3, q_4)$.
Although position and orientation are considered as different tasks \cite{kendall2015posenet} which should have some weighting factor during the learning process, we noticed that normalizing $(x, y)$ with respect to the total area size eliminates the need for such weighting. Our loss function is $\ell_2$ for the position $(x, y)$ and $\ell_2$ for the orientation $(q_1, q_2, q_3, q_4)$.

In a set of preliminary experiments we found that ResNet50 had the combination of smallest network size in terms of parameters and the best training and testing results. Therefore, we report our experiments using only the ResNet50 architecture. We decided not to use transfer learning using pre-trained weights, since the networks we tested were trained with ImageNet, which contains real photographs. Our assumption is that the pre-trained models are tuned for texture information that is not available in lean images. Hence, in our experiments, we trained the CNNs from scratch (note that we did use transfer learning within our setup; see Section~\ref{subsec:transfer_learning}).

\section{Tasks \& Hypothesis}
\label{sec:tasks_and_hypothesis}
We considered two localization tasks: retrieving the camera pose of an image from the training set (geo-matching), and recovering the camera pose of an image not in the training set (geo-interpolation).

Our goal was to answer the following questions: (i) Does geometry play a role when training the CNN for localization? (ii) Can a CNN be trained to solve these localization tasks from lean images?

\subsection{Geo-Matching Task}
\label{subsec:geo_matching_task}
Given an image from the \textbf{training~set}, we tested whether the correct camera pose could be determined. In a sense, the network is trained to overfit.
However, this would mean that the network managed to encode the entire set of images in some feature space as well as compute an efficient matching function between the features of the images to find the right pose.

\vspace{-0.3cm}\paragraph {(A) Geometrically Correlated:}
We examined whether a CNN can solve the geo-matching task using lean images. In this test, the camera pose for generating the image was used as ground truth for training. Hence, the pose of nearby images is correlated and the  network has access to this geometric information.




\vspace{-0.3cm}
\paragraph {(B) Geometrically Decorrelated:}
An alternative interpretation of the geo-matching task is that the network solves a simple indexing task, where the image's pose serves as a 4D label. 
Under this interpretation, the CNN does not use the available geometric information. Hence, an arbitrary labels of images would work just as well as in task A.
To test this, we used arbitrary poses as the image ground truth for task B. We randomly shuffled the pose information between images so that poses were not spatially correlated with respect to the images. 
If no geometry is used by the CNN, the results on this training data are expected to be similar to those obtained with the real pose as a ground truth.

\begin{figure}[h]
  \centering
  \begin{subfigure}[b]{0.45\linewidth}
    \centering
    \includegraphics[width=\textwidth]{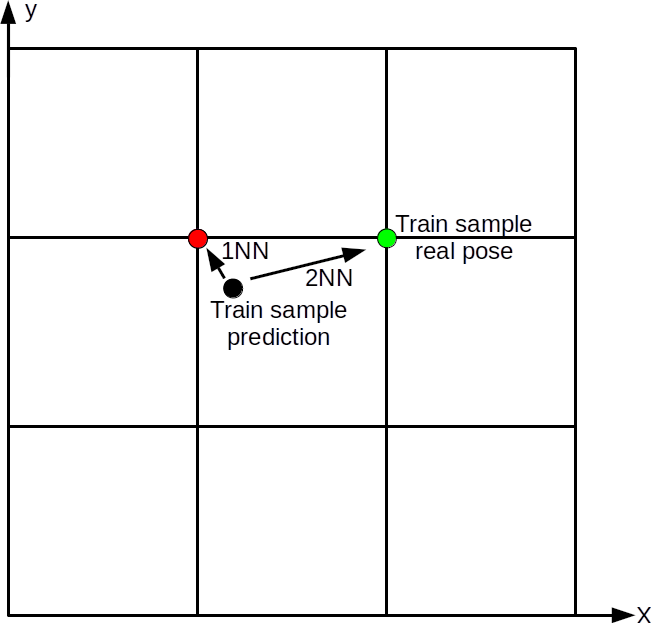}
    \caption{Nearest Neighbor}
    \label{subfig:distance_measures_nn}
  \end{subfigure}
  \begin{subfigure}[b]{0.45\linewidth}
    \centering
    \includegraphics[width=\textwidth]{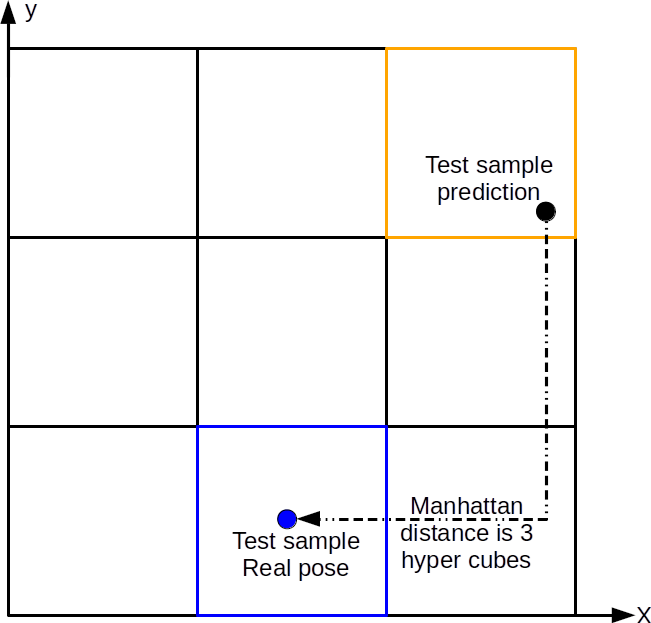}
    \caption{Manhattan Distance}
    \label{subfig:distance_measures_manhattan_distance}
  \end{subfigure}
  \caption{Illustration in 2D of the evaluation measures for geo-matching (\protect\subref{subfig:distance_measures_nn}) and geo-interpolation (\protect\subref{subfig:distance_measures_manhattan_distance}). The real measures are 4D in nature.}
  \label{fig:distance_measures}
\end{figure}

\vspace{-0.3cm}\paragraph{Evaluation:}
Since our network is a regression network, the computed pose does not necessarily match exactly a pose of an image from the training set (see Figure~\ref{fig:distance_measures}\subref{subfig:distance_measures_nn}). We used the nearest neighbor (nn) grid sample to the computed pose as the pose retrieval. We report the percentage of images whose correct pose is the nearest neighbor (1nn) and also report the percentage of images whose correct pose is among the three nearest neighbors (3nn) of the computed pose. These evaluations were used for both (A) and (B) geo-localization tasks. An additional advantage of using this measure is that it is given in grid steps and not in meter/angle units, circumventing the difficulty of comparing distances and angles and enabling a comparison of results with different grid densities (we do provide numerical $\ell_2$ errors in Table~\ref{tbl:l2_errors}).

\subsection{Geo-Interpolation Task (C)}
\label{sec:localization-task}
We tested whether the network can estimate the pose of an image that is not in the training set.
To avoid over-fitting and allow generalization, the network was trained until best result was achieved on a validation set.
We do not expect the network to return a correct position that is outside the learned area. Thus, this task is viewed as an interpolation task from known samples on the grid to unknown positions. For this reason our test set is comprised of images sampled at midpoints of the training grid. These are images that are farthest from the training set samples.

\vspace{-0.2cm}\paragraph{Evaluation:} 
We considered the hyper-cube of the computed pose.
A computed pose is considered correct if it lies within the same grid hyper-cube as the test sample.
We report the number of correctly computed poses ($D<1$). In addition, we considered the Manhattan distance between the hyper-cubes of the computed pose and the test sample (see Figure~\ref{fig:distance_measures}\subref{subfig:distance_measures_manhattan_distance}). We report the number of images for which this distance is smaller than~3 ($D<3$).
Note that these measurements are invariant to the sampling step size. Thus, we are able to compare results of experiments that were sampled with different step sizes.
For completness, we also provide the standard  $\ell_2$ errors in Table~\ref{tbl:l2_errors}.

\begin{table*}[h]
\centering
\begin{tabular}{|c|c||c|c|c||c|c|c|c|}
\hline

\multirow{4}{*}{} & \multirow{4}{*}{Input type} & \multicolumn{3}{c||}{Geo-Matching} & \multicolumn{4}{c|}{Geo-Interpolation} \\
\cline{3-5}
 & & (B) Arbitrary Pose &
 \multicolumn{2}{c||}{(A) Correct Pose} &
 \multicolumn{4}{c|}{(C) Correct Pose} \\
 \cline{3-9}
 & & \multirow{2}{*}{1nn} & \multirow{2}{*}{1nn} & \multirow{2}{*}{3nn} & \multicolumn{2}{c|}{2D $(x, y)$} & \multicolumn{2}{c|}{4D $(x, y, \theta, \phi)$} \\
& & & & & D\textless{}1 & D\textless{}3 & D\textless{}1 & D\textless{}3 \\
\hline

\multirow{3}{*}{\begin{tabular}[c]{@{}c@{}}
Area 400x400 \\ step 20 \\ 37K images \end{tabular}}
 & Edges & 0.45 & 0.97 & 0.99 & 0.64 & 0.82 & 0.58 & 0.75 \\
 & Faces & 0.35 & 0.99 & 0.99 & 0.56 & 0.76 & 0.51 & 0.69 \\
 & Depth & 0.23 & 0.99 & 0.99 & 0.61 & 0.79 & 0.55 & 0.72 \\
 & Edges+Faces & 0.29 & 0.98 & 0.99 & 0.72 & 0.88 & 0.65 & 0.82 \\
 & Edges+Faces+Depth & 0.24 & 0.98 & 0.99 & 0.71 & 0.88 & 0.64 & 0.81 \\
\hline

\multirow{3}{*}{\begin{tabular}[c]{@{}c@{}}
Area 400x400 \\ step 10 \\ 140K images \end{tabular}}
 & Edges & 0.11 & 0.98 & 0.98 & 0.85 & 0.94 & 0.84 & 0.93 \\
 & Faces & 0.05 & 0.97 & 0.97 & 0.80 & 0.90 & 0.79 & 0.88 \\
 & Depth & 0.06 & 0.97 & 0.97 & 0.83 & 0.92 & 0.82 & 0.91 \\
 & Edges+Faces & 0.09 & 0.97 & 0.97 & 0.88 & 0.96 & 0.87 & 0.95 \\
 & Edges+Faces+Depth & 0.08 & 0.94 & 0.95 & 0.88 & 0.96 & 0.87 & 0.95 \\
\hline

\multirow{3}{*}{\begin{tabular}[c]{@{}c@{}}
Area 800x800 \\ step 20 \\ 170K images \end{tabular}}
 & Edges & 0.06 & 0.96 & 0.96 & 0.62 & 0.78 & 0.59 & 0.75 \\
 & Faces & 0.01 & 0.96 & 0.96 & 0.51 & 0.68 & 0.48 & 0.65 \\
 & Depth & 0.01 & 0.96 & 0.97 & 0.61 & 0.77 & 0.59 & 0.73 \\
 & Edges+Faces & 0.04 & 0.92 & 0.93 & 0.70 & 0.86 & 0.67 & 0.83 \\
 & Edges+Faces+Depth & 0.03 & 0.95 & 0.96 & 0.70 & 0.85 & 0.67 & 0.81 \\
\hline
\end{tabular}

\caption{Results of our experiments.
The fraction of images on which a correct estimation was obtained out of the total number of valid images evaluated (the higher the better). For geo-matching we use the nearest neighbor measure (nn) and for geo-interpolation the Manhattan distance (D). Number of images -- average number of valid views in the training set of three experiments on different AOIs. See detailed discussion in the text.}
\label{tbl:results}
\vspace{-4mm} 
\end{table*}

\section{Experiments \& Results}
\label{sec:experiments_and_results}
We tested and evaluated the ResNet50 network for the three tasks described in Section~\ref{sec:tasks_and_hypothesis}. 
The datasets, which are described in Section~\ref{sec:data}, are defined by the following parameters:
\begin{enumerate}
  \item Area of interest (AOI): $(x, y, width, height)$.
  \item Grid-step, $\delta$: the distance between adjacent $(x,y)$ position of the sampling grid.
  That is, adjacent to $(x,y,\theta, \phi)$ are $(x\pm \delta,y,\theta, \phi)$ and $(x,y\pm \delta,\theta, \phi)$. The grid density in $(\theta, \phi)$ domain was fixed.
  \item Input type: edges, faces, depth, edges + faces, edges + faces + depth. For the last two input types the images were fed to the network by stacking them channel-wise.
  \item Validation set created by randomly choosing 10\% of the training samples.
  \item Test set created by images that were sampled at midpoints of the training grid.
\end{enumerate}

\begin{table}[b!]
\centering
\footnotesize
\tabcolsep=0.06cm
\begin{tabular}{|c||c|c|c|c||c|c|c|c|}
\hline

\multirow{4}{*}{} & \multicolumn{4}{c||}{(A) Geo-Matching} & \multicolumn{4}{c|}{(C) Geo-Interpolation} \\
 \cline{2-9}
 & \multicolumn{2}{c|}{$(x, y)$} & \multicolumn{2}{c||}{$(\theta, \phi)$} & \multicolumn{2}{c|}{$(x, y)$} & \multicolumn{2}{c|}{$(\theta, \phi)$} \\
& mean & median & mean & median & mean & median & mean & median \\
\hline

\multirow{3}{*}{\begin{tabular}[c]{@{}c@{}}
Area 400x400 \\ step 20 \\ 37K images \end{tabular}}
 & & & & & & & & \\
 & 3.65 & 3.26 & 0.84 & 0.69 & 26.30 & 11.26 & 10.95 & 1.84 \\
 & & & & & & & & \\
\hline

\multirow{3}{*}{\begin{tabular}[c]{@{}c@{}}
Area 400x400 \\ step 10 \\ 140K images \end{tabular}}
 & & & & & & & & \\
 & 2.37 & 2.10 & 0.57 & 0.48 & 7.99 & 3.67 & 2.65 & 0.67 \\
 & & & & & & & & \\
\hline

\multirow{3}{*}{\begin{tabular}[c]{@{}c@{}}
Area 800x800 \\ step 20 \\ 170K images \end{tabular}}
 & & & & & & & & \\
 & 5.43 & 4.71 & 0.67 & 0.54 & 40.23 & 12.28 & 9.80 & 1.40 \\
 & & & & & & & & \\
\hline

\end{tabular}

\caption{Examples of the $\ell_2$ errors for an experiment with Edges+Faces image types in each sub-space: spatial $(x,y)$ errors in (approx.) meters, and orientation $(\theta, \phi)$) errors in degrees. Similar to this example, usually the errors show a long-tail distribution: many images have small errors and a few have very large errors.}
\label{tbl:l2_errors}
\end{table}

We used various step sizes for the camera position on the grid: $\delta = 10, 20, 40$ in model units ($1$ unit$~\sim~1$ meter).
$\theta$~($yaw$) was sampled at $5^{\circ}$ steps between $0^{\circ}$ and $360^{\circ}$, and $\phi$ ($pitch$) was sampled at $3^{\circ}$ steps between $0^{\circ}$ and $15^{\circ}$.
The height was set to a fixed value of $z\simeq1.7$ above ground (human height) and no $roll$ was used.

We report the main results in Table~\ref{tbl:results}, for tasks (A)-(C). Each block of three rows consists of a different dataset, defined by the area size and $\delta$. For each block we considered the different types of lean images and evaluated them on the three tasks as described in Section~\ref{sec:tasks_and_hypothesis}. Each entry is an average of three different AOIs.
For completeness, Table~\ref{tbl:l2_errors} shows an example of the mean and median $\ell_2$ errors of the pose estimation for edges+faces experiment. Similar results were obtained in other experiments.
Table~\ref{tbl:low_grid_density_results} and Table~\ref{tbl:stress_test_results} show the results of testing the limitations of the CNN with respect to the sparsity of the grid ($\delta = 40$) and the size of the datasets ($>630K$ images).
We next discuss the obtained results.

\subsection{Geo-Matching}
Very poor results were obtained for the geo-matching task when arbitrary poses were used as ground truth (Table~\ref{tbl:results}--Task (B)), \ie when no geometric correlation between the images and their ground truth was available.
The highest percentage of correct matches ($45\%$) was obtained for the smallest set of considered images ($37K$ images).
For the largest set ($170K$ images), the percentage of correct matches was less than $10\%$.
As can be seen, the quality of the results decreases as the number training samples increases.
This is expected because for a fixed number of network parameters, memorization becomes impossible when more and more training samples are added. Note that we do not report on the 3nn measure, since it is meaningless for this randomized pose task.

\begin{table}[b]
\centering
\tabcolsep=0.05cm
\begin{tabular}{|c||c|c||c|c|c|c|}
\hline

\multirow{4}{*}{} & \multicolumn{2}{c||}{(A)} & \multicolumn{4}{c|}{(C)} \\

 &  \multicolumn{2}{c||}{ Geo-Matching} & \multicolumn{4}{c|}{Geo-Interpolation} \\
 
 \cline{2-7}
 & \multirow{2}{*}{1nn} & \multirow{2}{*}{3nn} & \multicolumn{2}{c|}{2D $(x, y)$} & \multicolumn{2}{c|}{4D $(x, y, \theta, \phi)$} \\
& & & D\textless{}1 & D\textless{}3 & D\textless{}1 & D\textless{}3 \\
\hline

\multirow{3}{*}{\begin{tabular}[c]{@{}c@{}}
Area 800x800 \\ step 10 \\ 636K images \end{tabular}}
 & & & & & & \\
 & 0.82 & 0.82 & 0.80 & 0.92 & 0.79 & 0.92 \\
 & & & & & & \\
\hline

\multirow{3}{*}{\begin{tabular}[c]{@{}c@{}}
Area 1600x1600 \\ step 20 \\ 666K images \end{tabular}}
 & & & & & & \\
 & 0.58 & 0.59 & 0.46 & 0.69 & 0.44 & 0.67 \\
 & & & & & & \\
\hline

\end{tabular}

\caption{Testing network learning capacity. These results are from a single experiment where the image input type is only edges. The network ability to learn drops when the number of images grows beyond a certain point.}
\label{tbl:stress_test_results}
\end{table}

\begin{table*}[h]
\centering
\begin{tabular}{|c|c||c|c||c|c|c|c|}
\hline

\multirow{4}{*}{} & \multirow{4}{*}{Input type} & \multicolumn{2}{c||}{(A) Geo-Matching} & \multicolumn{4}{c|}{(C) Geo-Interpolation} \\
 \cline{3-8}
 & & \multirow{2}{*}{1nn} & \multirow{2}{*}{3nn} & \multicolumn{2}{c|}{2D $(x, y)$} & \multicolumn{2}{c|}{4D $(x, y, \theta, \phi)$} \\
& & & & D\textless{}1 & D\textless{}3 & D\textless{}1 & D\textless{}3 \\
\hline

\multirow{3}{*}{\begin{tabular}[c]{@{}c@{}}
Area 800x800 \\ step 40 \\ 61K images \end{tabular}}
 & Edges & 0.90 & 0.96 & 0.39 & 0.62 & 0.30 & 0.49 \\
 & Edges+Faces & 0.91 & 0.97 & 0.48 & 0.72 & 0.38 & 0.59 \\
 & Edges+Faces+Depth & 0.96 & 0.98 & 0.48 & 0.72 & 0.38 & 0.58 \\
\hline

\multirow{3}{*}{\begin{tabular}[c]{@{}c@{}}
Area 1600x1600 \\ step 40 \\ 174K images \end{tabular}}
 & Edges & 0.89 & 0.90 & 0.22 & 0.39 & 0.19 & 0.32 \\
 & Edges+Faces & 0.94 & 0.95 & 0.30 & 0.50 & 0.24 & 0.41 \\
 & Edges+Faces+Depth & 0.94 & 0.96 & 0.32 & 0.51 & 0.26 & 0.43 \\
\hline

\multirow{3}{*}{\begin{tabular}[c]{@{}c@{}}
Area 800x800 \\ step 40 / sparse ang. \\ 2.5K images \end{tabular}}
 & Edges & 0.40 & 0.41 & \multicolumn{4}{c|}{} \\
 & Edges+Faces & 0.37 & 0.38 & \multicolumn{4}{c|}{Failed} \\
 & Edges+Faces+Depth & 0.26 & 0.27 & \multicolumn{4}{c|}{} \\
\hline

\multirow{3}{*}{\begin{tabular}[c]{@{}c@{}}
Area 1600x1600 \\ step 40 / sparse ang. \\ 7K images \end{tabular}}
 & Edges & 0.16 & 0.18 & \multicolumn{4}{c|}{} \\
 & Edges+Faces & 0.17 & 0.19 & \multicolumn{4}{c|}{Failed} \\
 & Edges+Faces+Depth & 0.13 & 0.14 & \multicolumn{4}{c|}{} \\
\hline

\end{tabular}

\caption{Low grid density results. Datasets (single experiment each) with sparser spatial sampling (top two blocks), and sparser spatial and orientation sampling (bottom two blocks) where the pitch angle is $\in[6, 12]$, and yaw $\in\{45i\}_{i=0}^7$.
The sparser the data, the worse the results. Geo-interpolation could not succeed in very sparse and very small datasets.}
\label{tbl:low_grid_density_results}

\vspace{-5mm}
\end{table*}

In contrast, when the correct poses were used as ground truth (Table~\ref{tbl:results}--Task (A)), 
the CNN succeeded in 1nn localization of more than $92\%$ of the training samples in all cases.
These results show that a CNN with around 8.5 million parameters is able to exploit the geometric structure of the scene and match an image with $\sim\!170K$ images with accuracy of~\mbox{$\sim90\%$}. That is, an average of $42.5$ parameters are used per image for images of size $160\times120=19200$ pixels.
Our interpretation is that using a CNN makes it possible to avoid the direct storage of the images (or its edges) and their labels. Given the trained network, the matching is much faster\footnote{Evaluation of an image in a batch takes $\sim1.7ms$ on Nvidia GeForce GTX 1080 GPU.} than with any traditional search algorithm on such a large dataset of images. 

We believe the significant differences between the two geo-matching tasks (A) and (B) is due to the network exploiting the geometric correlations when learning a metric between images. 

We also considered much larger datasets with more than $600K$ images (Table~\ref{tbl:stress_test_results}). The percentage of correct matches dropped to $82\%$ for a dense grid, $\delta=10$, and to $56\%$ for a sparser grid, $\delta=20$. For $\delta=20$ and $>600K$ images, the network capacity is probably saturated.
A comparison of these results to those reported in Table~\ref{tbl:results} (Task (A)) for the same $\delta$ values, indicates that both the number of images and the grid size determine how successfully the CNN models the data. 

In addition, we tested datasets with sparser sampling  in the position domain (Table~\ref{tbl:low_grid_density_results} top 2~blocks), and in both the position and the orientation domains (Table~\ref{tbl:low_grid_density_results} bottom 2~blocks).
For sparse sampling only in the position domain, the percentage of correct matches is reduced marginally. However, when reducing the sampling also in the orientation domain, the percentage of correct matches is dramatically dropped.
This indicates that 
it is easier for the CNN to model a denser grid (probably because of higher geometric correlation between images), and it is easier to model fewer images (probably because of network capacity).

\subsection{Geo-Interpolation}
\label{subsec:geo_interpolation}
Here we tested whether the pose estimation by the CNN generalizes to unseen images. We used the same training as in geo-matching with the correct pose as a ground truth, and we tested it on images sampled from the mid point of each grid cell. 
We report our results with respect to the 2D position as well as with respect to the 4D parameters of a pose (Table~\ref{tbl:results}). 

The network was able to generalize image position with good accuracy where $\sim70\%$ of images are correctly positioned in their grid cell, and above $80\%$ of the computed poses are within three cells of the correct one. As expected, this task achieves better results on a tighter grid ($\delta=10, \sim88\%$) than on a sparse grid ($\delta=20, \sim70\%$). 
The 4D position error is lower bounded by the 2D position error, and hence is greater. Moreover, the sampling rate in the orientation domain is much higher that in the location. Hence a small error in orientation estimation has a greater effect on the 4D errors. Still, the accuracy in 4D for $\delta=10$ is $\sim87\%$. 

We further tested the effect of the grid density. It is clear from Table~\ref{tbl:results}--Task(C) that for $\delta=10$ the results are better than for $\delta=20$, even if the number of images is larger. We further explore this observation for a sparser grid, $\delta=40$, where the percentage of correct estimations dropped significantly below $50\%$ and $30\%$ for $61K$ and $174K$ images, respectively (Table~\ref{tbl:low_grid_density_results}-Task(C)). For $\delta=10$ for $636K$ images, $80\%$ of the estimations were correct (Table~\ref{tbl:stress_test_results}-Task(C).
For this task, sparser sampling is more critical than the geo-matching task as can be seen in Table~\ref{tbl:low_grid_density_results}. For very sparse sampling of the $4D$ space the network cannot really generalize to positions not seen before. Here again
we believe that not only the number of images play a role but also their density. The denser the grid, the higher the correlation between images, and hence better generalization can be obtained.
 
A nice application of our results is the ability to rate the distinctiveness of positions in the city. In Figure~\ref{fig:input_images_and_test_vis} (bottom) we illustrate how certain places can be easily recognized (high geo-interploation success rate) while other are more difficult. Note, for instance, how open spaces are more distinct than narrow streets.

\subsection{Effect of Data Type}
\label{effect_of_data_type}
We compared the results on several types of lean images separately, and their combination. Faces alone provides the least geometric information, and indeed in most cases was inferior to edges or depth results. Surprisingly, edges alone provides better information than depth alone.

When combining edges with faces or with faces+depth, we expect the results of all tasks to be improved with respect to the results obtained when edges alone are used.
For the geo-matching task (A) with $\delta \leq 20$ (Table~\ref{tbl:results}), similar results were obtained for all data types. However, for a very sparse grid  (Table~\ref{tbl:low_grid_density_results} 2-upper blocks), richer geometric information improves the results. We believe it is because the results on $\delta \leq 20$ were very high to begin with with only edges.

\begin{figure}[b]
  \centering
  \includegraphics[width=1.0\linewidth]{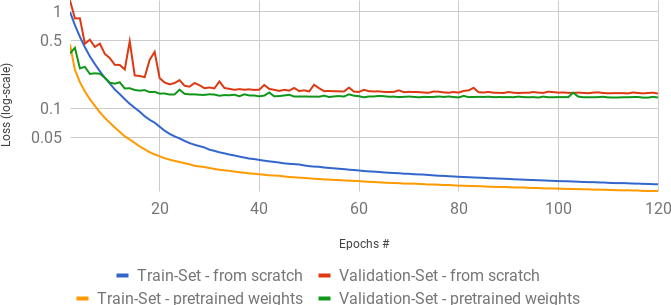}
  \caption{Transfer learning: learning from scratch vs. starting with pre-trained weights. These graphs are from one experiment where the input type was Edges+Faces, but similar behavior appeared in other experiments.}
  \label{fig:transfer_learning_1}
\end{figure}

For the geo-interpolation task (C), adding the faces information significantly improved the results, as expected. Surprisingly, the depth information did not show any significant performance gain when $\delta \leq 20$.
This may indicate that edges+faces provide sufficient information for these cases. However, for a very sparse grid, $\delta=40$, with a relatively small number of images, adding the depth significantly improves the results (Table~\ref{tbl:low_grid_density_results}, 174K images).

For the data with geometrically decorrelated pose (Task B) and for the very sparse sampling (Table~\ref{tbl:low_grid_density_results} bottom 2 blocks), the more information we add, the worse results were. The reason for this is still unclear to us. A possible explanation is that as the problem becomes more of a memorization task, the increase of information makes it harder for the CNN to find discriminant features.

\subsection{Transfer Learning}
\label{subsec:transfer_learning}
Once we had a trained a CNN for some AOI, we applied transfer learning to a new AOI by using the learned weights as initialization values for the new area. As can be seen in Figure~\ref{fig:transfer_learning_1}, doing so improved our learning rate. This indicates that the network managed to learn features of lean images that assist in other, similar experiments, and it does not depend only on memorization of the area for learning.

\section{Conclusions}
\label{sec:conclusions_and_future_work}
In this work we showed that (i) CNN can achieve good results in geo-localization tasks using only lean images taken from a very simple 3D model, and
(ii) that geometry plays an important role in geo-localization, by achieving good results while ignoring texture and scene details. The results indicates that noise-free lean images are sufficient for solving the geo-matching task using a CNN, and that the use of uncorrelated images makes it nearly impossible. In addition, our results indicate that (iii) geo-interpolation which is a generalization task, can also be solved by CNNs when using lean images.


From a more practical perspective, it is of interest to explore whether geometric information can be  used for real life geo-localization tasks, also because 3D models, \eg, the Open Street Map project~\cite{OpenStreetMap} are readily available.

{\small
\bibliographystyle{ieee}
\bibliography{references}
}

\end{document}